\documentclass{article}
\usepackage[preprint]{neurips_2026}
\usepackage{iftex}
\RequireXeTeX
\usepackage{fontspec}
\newfontfamily\zhfont{FandolSong-Regular.otf}[AutoFakeBold=2]
\XeTeXlinebreaklocale "zh"
\usepackage{url}
\usepackage{array}
\usepackage{booktabs}
\usepackage{float}
\usepackage{longtable}
\usepackage{colortbl}
\usepackage{tabularx}
\usepackage{amsmath}
\usepackage{amsfonts}
\usepackage{nicefrac}
\usepackage{microtype}
\usepackage{xcolor}
\usepackage{graphicx}
\usepackage{hyperref}

\newcommand{\keepaceneedspace}[1]{%
  \par
  \begingroup
  \dimen0=#1\relax
  \ifdim\pagegoal<\maxdimen
    \ifdim\dimexpr\pagegoal-\pagetotal\relax<\dimen0
      \newpage
    \fi
  \fi
  \endgroup
}

\title{Keepace.ai: A Technical Report on a Large Language Model for Sports and Health}

\author{%
  Keepace.ai Team \\
  \texttt{research@keep.com}
}

\begin{document}
\maketitle

\begin{abstract}
Large language models hold promise for delivering safe, scientifically grounded, and personalized exercise guidance at scale, yet the sports-and-health domain still lacks curated training corpora and rigorous evaluation benchmarks. Keepace.ai is a domain-specialized LLM built on Qwen3.5-27B through post-training, currently optimized for Chinese-language exercise guidance across three tasks: workout plan generation, sports-and-health knowledge QA, and personal sports data analysis. In internal rubric-based evaluations designed with domain experts, Keepace.ai outperforms its base model and larger general-purpose baselines, with the most pronounced improvements in exercise safety, boundary compliance, structured output completeness, and scientific accuracy. This report summarizes the current design, data pipeline, safety-first evaluation framework, and evaluation results of Keepace.ai.
\end{abstract}

\setcounter{section}{0}

\section{Introduction}

Physical inactivity ranks among the top modifiable risk factors for premature mortality. Globally, 31\% of adults and 80\% of adolescents fall short of recommended activity levels, with nearly 1.8 billion adults classified as inactive \citep{who2024physicalactivity}. Personalized exercise guidance can close this gap, yet qualified professionals are vastly outnumbered by demand, and exercise responses vary substantially across individuals \citep{exercise_variation2024}. Digital platforms have begun scaling access---Keep served approximately 29.9 million monthly active users in 2024 \citep{keep2024annual}---but at this volume, delivering safe, evidence-based, individualized assistance becomes a systems problem beyond human capacity alone. Large language models (LLMs), with their instruction-following and structured-generation capabilities, offer a natural automation path.

Applying LLMs to exercise guidance poses challenges distinct from clinical medicine \citep{singhal2023medpalm,singhal2025medpalm2}. Unlike clinical NLP, which benefits from benchmarks such as MedQA and MMLU \citep{medqa2021,mmlu2021}, exercise guidance lacks public benchmarks and instruction-tuning corpora; evaluation remains fragmented \citep{eval_health_coaches2025,exercise_llm_scoping2025}. The task family is also heterogeneous: workout planning demands structured artifacts with precise parameters; sports-and-health knowledge QA requires scientifically accurate explanation; personal sports data analysis calls for grounded reasoning over noisy physiological logs---a capability wearable-grounded health LLMs have only begun exploring \citep{healthllm2024,phllm2025,phia2026}. Finally, outputs must be personalized to diverse fitness levels, injury histories, and environments while never crossing into clinical diagnosis or treatment---a safety--personalization tension that general-purpose LLMs handle poorly.

We present \textbf{Keepace.ai}, a domain-specialized LLM for sports-and-health guidance built by post-training Qwen3.5-27B via supervised fine-tuning (SFT) and direct preference optimization (DPO). Keepace.ai is currently optimized and evaluated for Chinese-language sports-and-health guidance across three tasks---\emph{workout plan generation}, \emph{sports-and-health knowledge QA}, and \emph{personal sports data analysis}---unified by exercise safety and scientific accuracy. Development follows a closed-loop pipeline linking data construction, training, rubric-based evaluation, and case analysis. Task definitions and system scope are detailed in Section 2.

Main contributions:

\begin{itemize}
\item \textbf{Keepace.ai}, a sports-and-health LLM achieving the highest mean score among eight evaluated models (92.88), outperforming its base Qwen3.5-27B model (87.04) and the strongest comparison baseline, Qwen3.5-397B-A17B (88.53), with gains concentrated in exercise safety, boundary compliance, structured output completeness, and scientific accuracy in knowledge QA.
\item \textbf{A safety- and science-constrained data pipeline} converting heterogeneous inputs---in-house exercise courses, workout logs, and expert-curated sports-science literature---into SFT and DPO training data, filling the absence of public corpora for this domain.
\item \textbf{A rubric-based evaluation framework} designed by domain experts for the three tasks, covering exercise safety, scientific accuracy, coaching pedagogy, and personalization fidelity, applied via LLM-as-judge for scalable assessment.
\item \textbf{A two-stage SFT-then-DPO post-training procedure} with an auxiliary supervised objective during preference optimization to balance domain adaptation with general-capability preservation.
\end{itemize}

\begin{figure}[H]
\centering
\includegraphics[width=0.96\textwidth]{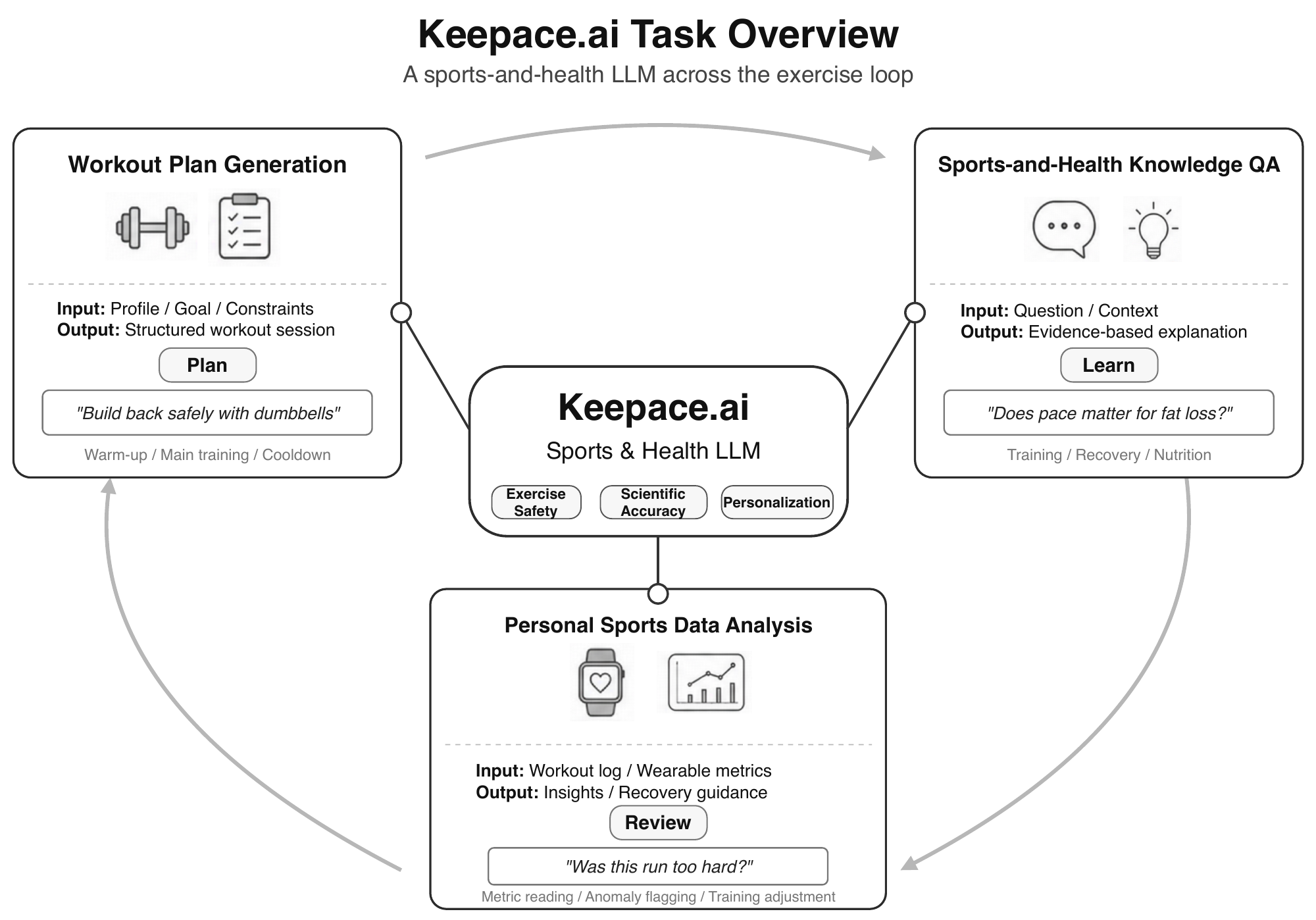}
\caption{\textbf{Keepace.ai task overview.} Keepace.ai is organized around an exercise-assistance loop: planning safe structured workouts, answering sports-science questions, and interpreting workout logs or wearable metrics for future adjustment.}
\label{fig:task_overview}
\label{fig:overview}
\label{fig:system_overview}
\end{figure}

\begin{figure}[H]
\centering
\includegraphics[width=0.96\textwidth]{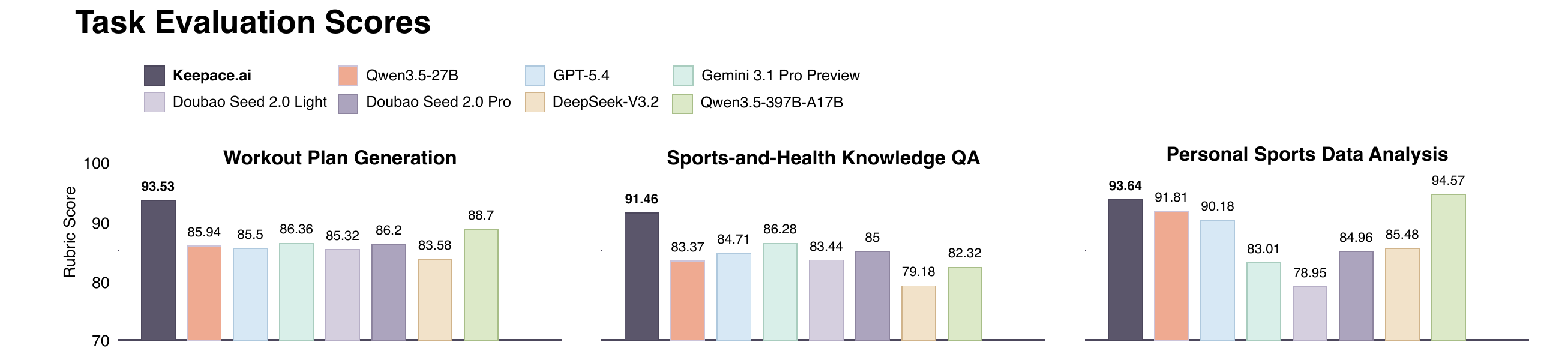}
\caption{\textbf{Keepace.ai evaluation summary.} Rubric-based evaluation scores for Keepace.ai and seven general-purpose baselines across workout plan generation, sports-and-health knowledge QA, and personal sports data analysis.}
\label{fig:evaluation_summary}
\label{fig:score_summary}
\end{figure}

Figure \ref{fig:task_overview} provides a compact overview of Keepace.ai's three target tasks across the exercise loop. Figure \ref{fig:evaluation_summary} summarizes its aggregate task-level performance against the same eight-model comparison set used in Section 6.

\setcounter{section}{1}

\section{Task definition}

\subsection{Scope and boundaries}

Keepace.ai is an exercise guidance assistant currently optimized and evaluated for Chinese-language sports-and-exercise guidance, covering workout plan generation, sports-and-health knowledge QA, and personal sports data analysis. It is explicitly \emph{not} intended for clinical diagnosis, rehabilitation prescription, medication guidance, or mental health treatment. Following established exercise-prescription safety logic \citep{chinese_exercise_prescription2023}, the model is trained to recommend professional consultation when user input suggests conditions outside the exercise-coaching scope, such as acute pain or cardiovascular warning signs.

\subsection{Task definitions}

The three tasks approximate a closed exercise-assistance loop: pre-exercise planning, coaching-oriented education, and post-exercise feedback. Table \ref{tab:task_definitions} summarizes task interfaces and scope constraints.

\begin{table}[H]
\caption{Task definitions and scope constraints for Keepace.}
\label{tab:task_definitions}
\centering
\begin{tabularx}{\textwidth}{>{\raggedright\arraybackslash}p{0.17\textwidth}>{\raggedright\arraybackslash}p{0.23\textwidth}>{\raggedright\arraybackslash}p{0.25\textwidth}>{\raggedright\arraybackslash}X}
\toprule
Task & Input & Output & Scope and quality focus \\
\midrule
Workout plan generation & User profile, training goal, and practical constraints, including fitness level, training history, reported limitations, equipment, space, and time budget. & Structured session with warm-up, main training, cooldown, exercise parameters, rest intervals, and accompanying session metadata. & Safety-first personalization; non-clinical exercise guidance; completeness and fit to user context. \\
\midrule
Sports-and-health knowledge QA & User question about exercise science, nutrition, recovery, or related topics, optionally with user profile context. & Educational answer with accurate terminology, evidence-grounded explanation, and explanation depth adapted to the user. & Coaching-oriented education; no individualized prescription or medical advice; scientific accuracy and usefulness. \\
\midrule
Personal sports data analysis & Structured exercise log and user query about the session or training pattern. & Evidence-grounded analysis that cites input metrics, flags anomalies, computes supported derived indicators, and recommends recovery or training adjustments. & Contextual interpretation of noisy wearable and mobile data; no diagnosis; accurate data use and actionable feedback. \\
\bottomrule
\end{tabularx}
\end{table}

\paragraph{Workout plan generation.}

Beyond the interface in Table \ref{tab:task_definitions}, the output includes session metadata: title, difficulty label, estimated duration, projected calorie expenditure, a course description summarizing objectives, preparation notes, and contraindication warnings tailored to the user's conditions. Completeness, accuracy, and readability are central quality dimensions for AI-generated exercise recommendations \citep{exercise_recommendations_chatbot2024}.

\paragraph{Sports-and-health knowledge QA.}

The model produces coaching-oriented education rather than fact retrieval: it explains principles and provides evidence-grounded guidance without individualized prescriptions or medical recommendations. We adopt this framing following evidence that behavior-science approaches improve actionability and empathy in activity-coaching outputs \citep{behavior_science_llm_coach2024}.

\paragraph{Personal sports data analysis.}

The model identifies sensor anomalies, explains observed patterns mechanistically (e.g., heart-rate drift as an aerobic-ceiling indicator), computes derived indicators (e.g., estimated maximum heart rate, training-load trends), and correlates across metrics. Because wearable data are noisy, heterogeneous, and device-dependent, the task emphasizes contextual interpretation over mechanical reading of raw values \citep{wearable_sleep_circadian2024,phia2026}.

These task definitions establish the interfaces used for data construction, post-training, and evaluation throughout the report.

\setcounter{section}{2}

\section{Training data creation}

Our central design principle is that safety constraints and scientific grounding should be targeted at the data level, rather than left to the model to learn implicitly from noisy examples.

\subsection{Data challenges}

Table \ref{tab:data_risks} summarizes five categories of data risk and their mitigations.

\begin{table}[H]
\caption{Data risks and mitigation strategies in training-data construction.}
\label{tab:data_risks}
\centering
\begin{tabularx}{\textwidth}{>{\raggedright\arraybackslash}p{0.22\textwidth}>{\raggedright\arraybackslash}X>{\raggedright\arraybackslash}p{0.24\textwidth}}
\toprule
Risk category & Example & Mitigation approach \\
\midrule
Fitness pseudoscience & "Sweating more means burning more fat" & Expert-knowledge filtering \\
Medical overreach & Generating clinical diagnoses or rehabilitation prescriptions & Domain-boundary filtering \\
Unsafe exercise programming & Excessive plyometrics for novice users & Rule-based safety constraints \\
Privacy exposure & Personally identifiable information in exercise logs & Anonymization \\
Sensor anomaly & Physiologically implausible elevation or pace readings & Anomaly flagging \\
\bottomrule
\end{tabularx}
\end{table}

The first three categories directly affect user safety and constitute the primary design targets of the data pipeline; the remaining two are data-quality prerequisites resolved before structured records become training targets.

\subsection{Data sources}

Training data are drawn from two complementary categories; all platform data are anonymized before entering the pipeline.

\paragraph{In-house exercise data from the Keep platform}

includes structured workout logs with physiological signals (heart rate, training load, RPE), professionally designed exercise courses with movement specifications, and user profiles capturing training history and body metrics. Before these platform-derived records are used for training-data construction, they undergo privacy-preserving de-identification to prevent leakage of user-level private information; Section 3.3 details the governance and filtering process. Course specifications and user profiles drive workout plan generation; workout logs ground personal sports data analysis; and user profiles contribute personalization context across all tasks. Recent wearable-grounded LLM work confirms that such records can support health reasoning while requiring careful transformation for language-model training \citep{healthllm2024,phllm2025,sensorlm2025}.

\paragraph{Domain literature curated by an internal sports-science expert team}

includes exercise physiology textbooks, training methodology manuals, nutrition references, and internally developed coaching standards---thousands of curated documents in total. These serve as the primary knowledge source for sports-and-health knowledge QA and provide scientific grounding for safety constraints. Domain experts review all materials to ensure the knowledge base reflects established exercise-science evidence.

\subsection{Data governance and privacy}

Before platform-derived records are used for training, they are de-identified and converted into structured or aggregated representations. The training pipeline does not use direct identifiers or other personally identifiable information (PII); user logs enter the process as task-relevant exercise summaries rather than raw identity-linked histories. Sensitive fields, residual PII, and corrupted records are filtered before example construction, and data handling follows internal privacy and security governance procedures. These controls align the training objective with sports-and-exercise guidance while excluding uses aimed at medical diagnosis, user identification, or identity-level profiling.

\subsection{Data processing pipeline}

\begin{figure}[H]
\centering
\includegraphics[width=0.96\textwidth]{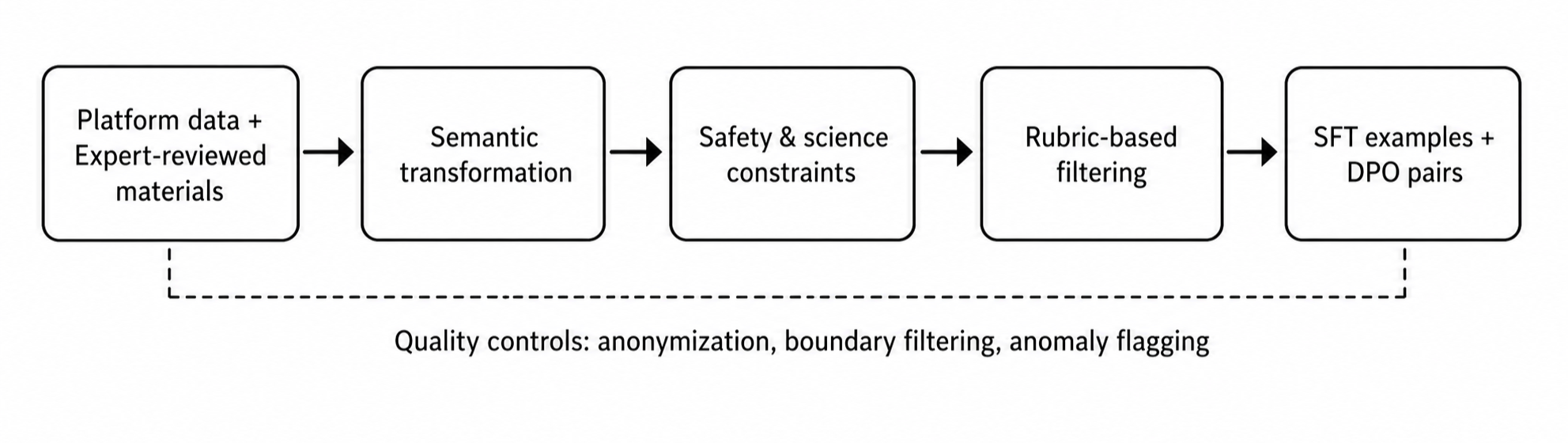}
\caption{\textbf{Training data creation pipeline.} Platform records and expert-reviewed materials are converted into training examples through semantic transformation, safety- and science-constraint injection, and rubric-based quality filtering. DPO preference pairs are constructed on top of the same supervised data infrastructure.}
\label{fig:keepace_training_data_pipeline}
\end{figure}

Figure \ref{fig:keepace_training_data_pipeline} summarizes the end-to-end process. Raw sources pass through three stages designed to embed safety constraints and scientific grounding into the training data.

\subsubsection{Stage 1: Semantic transformation of structured data}

Structured platform records must be abstracted into natural-language representations that support language reasoning \citep{sensorlm2025,mobile_behavioral_health_llm2024}.

\paragraph{Example A: Exercise log.}

A structured running log containing fields such as \texttt{duration: 3021s}, \texttt{avg\_hr: 158}, \texttt{max\_hr: 186}, \texttt{zone\_distribution: \{aerobic: 34\%, threshold: 7\%, anaerobic: 52\%, peak: 7\%\}}, \texttt{rpe: 7}, \texttt{training\_load: 218} is rendered as:

\begin{quote}
\emph{The user completed a 50-minute outdoor run. Average heart rate was 158 bpm (maximum 186 bpm), with 52\% of the session spent in the anaerobic endurance zone and 34\% in the aerobic power zone. Subjective exertion was rated 7 out of 10, and the estimated training load was 218.}
\end{quote}

\paragraph{Example B: Course specification.}

A structured course object listing \texttt{section: warmup, exercises: [\{name: "shoulder\_circles", duration: 20s\}, \{name: "hip\_mobility", duration: 20s\}], section: training, exercises: [\{name: "half\_squat", reps: 12, rest: 15s\}, \{name: "reverse\_lunge", reps: 10, per\_side: true, rest: 30s\}, \{name: "glute\_bridge", reps: 12, rest: 20s\}]} is rendered as:

\begin{quote}
\emph{Warm-up: shoulder circles (20 s), hip mobility drill (20 s). Main training: half squats (12 reps, 15 s rest), reverse lunges (10 reps per side, 30 s rest), and glute bridges (12 reps, 20 s rest), with coaching cues emphasizing controlled descent and stable knee tracking.}
\end{quote}

User profiles are similarly linearized into natural-language descriptions combining demographic data, fitness-level indicators, and recent training statistics.

\subsubsection{Stage 2: Safety and scientific constraint injection}

This stage addresses the data risks in Section 3.1 through three complementary mechanisms.

\paragraph{Rule-based safety constraints}

encode exercise-science norms as guardrails. For workout plan generation these include: movement-pattern load limits for risk-sensitive scenarios (e.g., tighter repetition caps for hip-hinge exercises; reduced plyometric volume for novices), mandatory rest intervals between exercises sharing the same movement pattern, and prohibition of high-impact movements late in sessions where fatigue compromises control. These rules derive from expert consensus and established exercise-prescription principles \citep{chinese_exercise_prescription2023,exercise_variation2024}.

\paragraph{Knowledge grounding}

ensures scientific content derives from expert-curated literature (Section 3.2) rather than unconstrained generation. Relevant source passages are provided as context during target construction, so explanations trace back to authoritative references rather than parametric knowledge alone.

\paragraph{Domain-boundary filtering}

removes content crossing into clinical diagnosis, pharmaceutical recommendation, or rehabilitation prescription, and targets common fitness pseudoscience.

\subsubsection{Stage 3: Rubric-based quality evaluation}

This stage applies the task-specific rubrics described in Section 5 as quality gates, following prior work showing that skill-level evaluation is more diagnostic than a single holistic score \citep{flask2024}. Only examples passing rubric-based evaluation enter the final training corpus; the same rubrics also serve as the quality criterion for preference data construction.

\subsection{Preference data construction}

We construct preference data for DPO on top of the SFT pipeline. For each prompt, candidates are generated by multiple models with complementary strengths; chosen--rejected pairs are assembled based on rubric scores, ensuring preference signal consistent with SFT quality standards.

The rubric is used in both data construction and final evaluation (Section 6). To mitigate circular optimization, test prompts are strictly held out from all data-construction stages; data construction operates on training-distribution prompts, whereas evaluation operates on held-out prompts.

The resulting corpus contains a large supervised set and a compact preference set covering all three target tasks, with exact mixture and scaling details treated as internal training configuration.

\setcounter{section}{3}

\section{Methods}

\subsection{Base model selection}

We select \textbf{Qwen3.5-27B} \citep{qwen35_2026,qwen35_27b_modelcard} as the base model because its open weights support controlled domain adaptation, internal safety evaluation, and production-oriented deployment. At 27 billion parameters, it provides a practical balance between instruction-following capability, multilingual reasoning, and the iteration speed required for closed-loop data--training--evaluation development.

\subsection{Post-training procedure}

We adapt the base model through two sequential stages: supervised fine-tuning (SFT), followed by direct preference optimization (DPO) \citep{dpo2023} with an auxiliary SFT objective. Both stages use the domain-specific data from Section 3.

\paragraph{Stage 1: Supervised fine-tuning.}

The base model is fine-tuned on the supervised corpus spanning all three tasks by minimizing the standard autoregressive negative log-likelihood:

\begin{equation}
\mathcal{L}_{\text{SFT}}(y \mid x) = -\sum_{t=1}^{|y|} \log p_\theta(y_t \mid x, y_{<t})
\end{equation}

where $x$ is the input context, $y$ is the target response, and $\theta$ denotes the model parameters.

\paragraph{Stage 2: Preference optimization with a joint objective.}

SFT produces outputs that follow the target format but may fall short on preference-sensitive dimensions such as personalization depth or pedagogical quality. Starting from the SFT checkpoint, we optimize on the preference data from Section 3.5. Given a preference pair $(y_w, y_l)$ for input $x$, the DPO loss is \citep{dpo2023}:

\begin{equation}
\mathcal{L}_{\text{DPO}}(y_w, y_l \mid x) = -\log \sigma \!\left( \beta \left[ \log \frac{p_\theta(y_w \mid x)}{p_{\text{ref}}(y_w \mid x)} - \log \frac{p_\theta(y_l \mid x)}{p_{\text{ref}}(y_l \mid x)} \right] \right)
\end{equation}

where $p_{\text{ref}}$ is a frozen reference model, $\sigma$ is the sigmoid function, and $\beta$ is a temperature parameter. In practice, the preference objective is optimized together with an auxiliary supervised likelihood term on the preferred response. This keeps the model anchored to the supervised task distribution while preference optimization shifts behavior toward safer, more personalized, and more rubric-aligned outputs \citep{orpo2024}. The frozen $p_{\text{ref}}$ provides a stable anchor throughout training. We examine this objective choice through a structured-generation case study in Section 6.2.

\subsection{Training details}

Both stages perform full-parameter fine-tuning on the 27B-parameter model. Table \ref{tab:training_config} summarizes the training configuration.

\begin{table}[H]
\caption{Training configuration for the SFT and DPO stages.}
\label{tab:training_config}
\centering
\begin{tabularx}{\textwidth}{>{\raggedright\arraybackslash}p{0.28\textwidth}>{\raggedright\arraybackslash}X>{\raggedright\arraybackslash}X}
\toprule
 & \textbf{SFT stage} & \textbf{DPO stage} \\
\midrule
Initialization & Qwen3.5-27B base weights & SFT checkpoint \\
Training data & ${\sim}$200K supervised examples & ${\sim}$4K preference pairs \\
Objective & $\mathcal{L}_{\text{SFT}}$ & $\mathcal{L}_{\text{DPO}} + \alpha\,\mathcal{L}_{\text{SFT}}$ \\
Learning rate & $1 \times 10^{-6}$ & $1 \times 10^{-6}$ \\
Batch size & 64 & 32 \\
$\alpha$ & --- & 0.7 \\
\bottomrule
\end{tabularx}
\end{table}

All training was conducted on the Alibaba Cloud PAI platform. The final model was selected based on rubric-evaluated performance on a held-out validation split distinct from the test sets in Section 6, ensuring no evaluation leakage. The DPO-stage configuration, including the auxiliary supervised weight $\alpha = 0.7$, was chosen via the same validation procedure.

\setcounter{section}{4}

\section{Evaluation framework}

The sports-and-health domain lacks widely adopted public benchmarks for exercise guidance systems \citep{eval_health_coaches2025,exercise_llm_scoping2025}. We therefore design a task-specific evaluation framework in collaboration with domain experts, organized around \textbf{rubrics}---atomic criteria each testing exactly one response aspect---and \textbf{dimensions}, thematic groupings of related rubrics.

\subsection{Design principles}

\paragraph{Flat, orthogonal, binary rubrics.}

Each rubric tests one aspect with ternary scores (1 pass / 0 fail / -1 not applicable), ensuring each deficiency is penalized exactly once and improving inter-rater agreement \citep{flask2024,scalable_health_eval2026,adaptive_boolean_exercise2025}.

\paragraph{Evidence-grounded judgment.}

Every score must be accompanied by a justification citing specific response content, improving transparency for both human annotators and LLM judges \citep{geval2023,closerlook2023}.

\paragraph{Domain-expert design with cross-rubric coordination.}

Rubrics are designed by exercise-science experts with explicit coordination rules delineating scope to prevent overlapping assessments.

\subsection{Task-specific rubrics}

The framework instantiates these design principles separately for the three target tasks. Each task uses a dimension-level rubric table for reporting, with finer-grained criteria maintained internally for automated judging and expert auditing.

\subsubsection{Workout plan generation}

The workout plan evaluation comprises 35 rubrics in 8 dimensions (Table \ref{tab:rubric_workout}), reflecting exercise-prescription principles \citep{chinese_exercise_prescription2023,exercise_recommendations_chatbot2024}.

\begin{table}[H]
\caption{Overview of the workout plan generation rubric.}
\label{tab:rubric_workout}
\centering
\begin{tabularx}{\textwidth}{llc>{\raggedright\arraybackslash}X}
\toprule
Dim. & Dimension Name & \# Rubrics & Scope \\
\midrule
A & Exercise Safety & 2 & Acute risk recognition (halting on dangerous symptoms); professional scope compliance (not overstepping into clinical diagnosis or rehabilitation) \\
B & Course Metadata & 4 & Title, difficulty label, duration estimate, calorie projection \\
C & Course Description & 5 & Training objective, structure overview, preparation notes, contraindications, expectation management \\
D & Content Readability & 4 & Section naming, exercise naming, parameter completeness, terminology accuracy \\
E & Course Adaptability & 4 & Goal alignment, programming style match, expected post-session feeling, overall load appropriateness \\
F & Environment Adaptation & 5 & Equipment, space, time budget, social setting, time-of-day constraints \\
G & Choreography Quality & 7 & Three-phase structure, warm-up targeting, cooldown targeting, rest intervals, exercise order, muscle-group balance, transition flow \\
H & Movement Adaptability & 4 & Executability for user level, injury/condition adaptation, difficulty distribution, inherent movement risk \\
\bottomrule
\end{tabularx}
\end{table}

The A/H separation distinguishes macro-level safety awareness (A: recognizing acute risk signals, staying within coaching scope) from micro-level movement selection (H: executability and adaptation for the specific user), so a plan can score well on one while failing the other. Choreography quality (G) is the largest dimension (7 rubrics), encoding exercise-science sequencing and rest-interval principles.

\subsubsection{Sports-and-health knowledge QA}

The sports-and-health knowledge QA evaluation comprises 14 rubrics in 4 dimensions.

\begin{table}[H]
\caption{Overview of the sports-and-health knowledge QA rubric.}
\label{tab:rubric_qa}
\centering
\begin{tabularx}{\textwidth}{llc>{\raggedright\arraybackslash}X}
\toprule
Dim. & Dimension Name & \# Rubrics & Scope \\
\midrule
A & Scientific Accuracy & 4 & Core factual correctness, terminology precision, evidence sourcing, unbiased comparison of training philosophies \\
B & Pedagogical Quality & 5 & Fitness-domain relevance, audience-level matching, myth debunking, analogy/visualization quality, actionable granularity \\
C & Expression Quality & 2 & Answer structure and logic, empathy and communication tone \\
D & Boundary \& Safety & 3 & Medical-boundary restraint, coaching-vs-prescription restraint, risk warnings for dangerous topics \\
\bottomrule
\end{tabularx}
\end{table}

Scientific accuracy (A) is the central concern in a domain where pseudoscientific claims are pervasive. Audience-level matching (B2) checks register adaptation to the user's knowledge level, while the coaching-versus-prescription boundary (D2) verifies that the model provides educational content rather than individualized prescriptions, reflecting the scope-of-practice principle from Section 3.1.

\subsubsection{Personal sports data analysis}

The personal sports data analysis evaluation comprises 17 rubrics in 6 dimensions.

\begin{table}[H]
\caption{Overview of the personal sports data analysis rubric.}
\label{tab:rubric_interp}
\centering
\begin{tabularx}{\textwidth}{llc>{\raggedright\arraybackslash}X}
\toprule
Dim. & Dimension Name & \# Rubrics & Scope \\
\midrule
A & Data Comprehension & 3 & Data citation accuracy, missing-data handling, anomaly identification \\
B & Analysis \& Interpretation & 4 & Metric coverage and goal relevance, mechanistic explanation, derived-indicator reasoning, multi-metric correlation \\
C & Expression Quality & 4 & Terminology accuracy, data contextualization, answer structure, empathy \\
D & Risk \& Recommendations & 2 & Risk-signal identification, recommendation safety \\
E & Recommendation Quality & 2 & Scientific basis, actionability and granularity matching \\
F & Medical Boundary & 2 & Acute-symptom handling, no clinical diagnosis from exercise data \\
\bottomrule
\end{tabularx}
\end{table}

\paragraph{Derived-indicator reasoning (B3)}

evaluates whether the model computes physiologically meaningful indicators beyond directly reported metrics (e.g., estimating max HR from age, computing heart-rate drift, projecting race times). \textbf{Multi-metric correlation (B4)} assesses whether the model integrates multiple data dimensions into synthetic conclusions rather than commenting on metrics in isolation; it is triggered when the log contains three or more dimensions or the user asks about overall performance.

\subsection{Scoring protocol}

\paragraph{Score computation.}

Each applicable rubric receives 1 (pass) or 0 (fail); rubrics scored -1 are excluded. The per-example score is:

\begin{equation}
S_i = \frac{\sum_{j \in \mathcal{A}_i} \mathbf{1}[s_{ij} = 1]}{|\mathcal{A}_i|} \times 100
\end{equation}

where $\mathcal{A}_i$ is the set of applicable rubrics for example $i$ and $s_{ij} \in \{0, 1\}$. The per-task score averages over all $N$ examples:

\begin{equation}
S_{\text{task}} = \frac{1}{N} \sum_{i=1}^{N} S_i
\end{equation}

\paragraph{Automated alignment and judging pipeline.}

To support repeatable model iteration, we implement the rubric protocol as a high-throughput automated evaluation pipeline. Claude Sonnet 4 \citep{claude4_systemcard2025} serves as the structured judge, following the rubric-based evaluation lineage of G-Eval, FLASK, and Prometheus \citep{geval2023,flask2024,prometheus2_2024}. The judge receives full rubric definitions and produces structured JSON with a score and evidence-based justification per rubric. This automated pipeline is paired with human-in-the-loop expert auditing for calibration and governance. Full internal rubric definitions and judge prompts are maintained for evaluation governance but are not released in this report; the summarized rubric structure remains in Tables 4--6.

\paragraph{Agreement with human experts.}

For each task, 30 responses sampled from multiple models are independently scored by three certified sports-science professionals using the same protocol, blind to model identity; disagreements are resolved by majority vote.

\begin{table}[H]
\caption{Agreement between the LLM-based judge and human expert annotations.}
\label{tab:judge_agreement}
\centering
\begin{tabular}{lccc}
\toprule
Task & \# Applicable Judgments & Agreement (\%) & Cohen's $\kappa$ \\
\midrule
Workout Plan Generation & 832 & 91.3 & 0.78 \\
Sports-and-Health Knowledge QA & 354 & 89.4 & 0.74 \\
Personal Sports Data Analysis & 418 & 90.7 & 0.76 \\
\textbf{Overall} & \textbf{1604} & \textbf{90.6} & \textbf{0.76} \\
\bottomrule
\end{tabular}
\end{table}

The LLM judge agrees with experts on over 89\% of rubric judgments ($\kappa$ 0.74--0.78, substantial agreement). Disagreements concentrate on rubrics requiring nuanced domain judgment rather than clear-cut criteria, indicating the judge is a useful proxy for expert evaluation at scale \citep{nofreelabels2025,rubrics_backfire2026}. We report rubric-level agreement because aggregate scores can mask failures on individual criteria \citep{rubriceval2026}.

\paragraph{Evaluation settings.}

The eight-model comparison uses an April 2026 internal evaluation snapshot with 100 held-out Chinese-language prompts per task. All compared models receive identical task prompts and system instructions, are judged with the same rubric definitions, structured JSON scoring schema, and Claude Sonnet 4 judge. Tool use and web search are disabled for every model. Reasoning mode is disabled. Temperature, maximum output tokens, and other decoding controls are matched across models. Because provider-side decoding internals may still differ, results should be interpreted as an internally controlled comparison under standardized prompting, judging, and output-budget conditions.

\subsection{Test set construction}

For each task, 100 held-out samples are drawn from the same prompt distribution as training data but excluded from all training stages. Three isolation constraints prevent leakage: user-level separation, near-duplicate filtering (n-gram overlap and semantic similarity), and temporal ordering. Samples are stratified for difficulty and sub-type coverage and reviewed by domain experts for realism. All models receive identical inputs under the same rubric definitions and judge configuration.

\setcounter{section}{5}

\section{Results}

\subsection{Main results}

We evaluate Keepace.ai against seven baselines summarized in Figure \ref{fig:evaluation_summary}: its base model Qwen3.5-27B \citep{qwen35_27b_modelcard}, a larger Qwen3.5-397B-A17B \citep{qwen35_397b_modelcard}, GPT-5.4 \citep{gpt54_2026}, Gemini 3.1 Pro Preview \citep{gemini31pro_2026}, DeepSeek-V3.2 \citep{deepseekv32_2025}, and two Doubao Seed 2.0 variants \citep{doubao2_2026}. All models are evaluated in non-thinking mode under the rubric-based framework described in Section 5. Table \ref{tab:main_results} reports per-task scores and the overall mean.

\begin{table}[H]
\caption{Main results across the three target tasks and the overall mean score.}
\label{tab:main_results}
\centering
\begin{tabular}{lcccc}
\toprule
Model & Workout Plan & Knowledge QA & Data Analysis & Mean \\
\midrule
\textbf{Keepace.ai} & \textbf{93.53} & \textbf{91.46} & 93.64 & \textbf{92.88} \\
Qwen3.5-397B-A17B & 88.70 & 82.32 & \textbf{94.57} & 88.53 \\
Qwen3.5-27B & 85.94 & 83.37 & 91.81 & 87.04 \\
GPT-5.4 & 85.50 & 84.71 & 90.18 & 86.80 \\
Doubao Seed 2.0 Pro & 86.20 & 85.00 & 84.96 & 85.39 \\
Gemini 3.1 Pro Preview & 86.36 & 86.28 & 83.01 & 85.22 \\
DeepSeek-V3.2 & 83.58 & 79.18 & 85.48 & 82.75 \\
Doubao Seed 2.0 Lite & 85.32 & 83.44 & 78.95 & 82.57 \\
\bottomrule
\end{tabular}
\end{table}

Keepace.ai achieves the highest overall mean score (92.88), outperforming the strongest baseline by +4.35 points (Qwen3.5-397B-A17B, 88.53) and its base model by +5.84 points (Qwen3.5-27B, 87.04). It ranks first on workout plan generation and sports-and-health knowledge QA, and second on personal sports data analysis behind Qwen3.5-397B-A17B.

\paragraph{Consistent improvement across tasks.}

Baselines exhibit uneven profiles: Qwen3.5-397B-A17B leads on data analysis but drops on knowledge QA, Gemini 3.1 Pro Preview performs strongly on knowledge QA but weakly on data analysis, and Doubao Seed 2.0 Lite drops sharply on data analysis. Keepace.ai maintains scores above 91 on every task, indicating broad rather than task-specific improvement.

\paragraph{Largest absolute gains on workout plan generation.}

Relative to its base model, Keepace.ai's largest task-level gain is on workout plan generation (+7.59), which carries the most stringent safety and structural requirements in our rubric. Compared with the strongest non-Keepace.ai model on this task, Qwen3.5-397B-A17B, Keepace.ai still leads by +4.83 points, suggesting that domain-specific data construction and constraint injection are particularly effective for structured, safety-sensitive generation.

\subsection{Targeted Ablation Study on Alignment Trade-offs}

To examine the training-objective choice introduced in Section 4.2, we conduct a targeted ablation study on alignment trade-offs using workout plan generation as a probe task. This task is deliberately chosen because it combines the three properties most central to Keepace.ai: structured output generation, personalized exercise constraints, and explicit safety-boundary requirements. The study isolates how different post-training objectives trade off domain-task improvement against general-capability preservation, complementing the full multi-task comparison in Table \ref{tab:main_results}.

Starting from the Qwen3.5-27B base model, we compare three training configurations: SFT only, DPO only, and the joint DPO objective with auxiliary supervised anchoring introduced in Section 4.2. Each configuration is evaluated on two general-capability benchmarks (CEval and IFBench) alongside the workout plan generation rubric. The workout-plan score is produced by the same continuous alignment evaluation pipeline described in Section 5.3: rubric-level automated judging with Claude Sonnet 4, evidence-grounded JSON scoring, and human-in-the-loop expert auditing. Table \ref{tab:objective_ablation} summarizes the comparison.

\begin{table}[H]
\caption{Targeted ablation study on alignment trade-offs using workout plan generation as the probe task.}
\label{tab:objective_ablation}
\centering
\begin{tabular}{lccc}
\toprule
Variant & CEval & IFBench & Workout Plan \\
\midrule
Qwen3.5-27B base & 90.79 & 44.33 & 85.94 \\
SFT only & 88.04 & 31.67 & \textbf{94.61} \\
DPO only & \textbf{91.08} & 46.33 & 86.18 \\
Joint SFT--DPO objective & 90.93 & \textbf{46.67} & 92.88 \\
\bottomrule
\end{tabular}
\end{table}

Pure SFT yields the highest domain-task score in this setting (94.61), but at a substantial cost to general capabilities: CEval drops to 88.04 and IFBench falls to 31.67. This indicates that aggressive supervised adaptation can overfit the model to domain-specific output formats while degrading general knowledge and instruction-following ability.

Pure DPO presents the converse trade-off. General-capability scores remain intact or slightly improve (CEval 91.08, IFBench 46.33), yet the workout plan score rises only marginally to 86.18. Preference optimization alone does not supply the structured domain knowledge that workout plan generation demands.

The joint objective achieves the most favorable balance: it raises the workout plan score to 92.88 while maintaining CEval at 90.93 and improving IFBench to 46.67. Although its domain-task gain is smaller than that of pure SFT, it avoids the associated general-capability degradation. Based on this comparison, we adopted the joint SFT--DPO objective for all subsequent multi-task training, leading to the final scores in Table \ref{tab:main_results}.

\subsection{Per-dimension gains over the base model}

Since Keepace.ai is obtained by post-training Qwen3.5-27B, this section compares Keepace.ai with its base model at the rubric-dimension level. The goal is not to rank external models again, but to identify which domain-specific capabilities are improved by the SFT and DPO pipeline. We therefore report base-model deltas for each task, using the rubric dimensions defined in Section 5.

\subsubsection{Workout plan generation}

Table \ref{tab:dim_workout} reports base-model deltas for workout plan generation (8 dimensions, 35 rubrics).

\begin{table}[H]
\caption{Per-dimension gains over the Qwen3.5-27B base model for workout plan generation.}
\label{tab:dim_workout}
\centering
\begin{tabular}{lccc}
\toprule
Dimension & Qwen3.5-27B & Keepace.ai & Delta \\
\midrule
A. Exercise Safety & 60.87 & \textbf{93.33} & \textbf{+32.46} \\
B. Course Metadata & 40.50 & \textbf{93.50} & \textbf{+53.00} \\
C. Course Description & 88.56 & \textbf{97.22} & \textbf{+8.66} \\
D. Content Readability & \textbf{96.50} & 96.00 & -0.50 \\
E. Course Adaptability & \textbf{98.00} & 94.00 & -4.00 \\
F. Environment Adaptation & 93.26 & \textbf{94.67} & \textbf{+1.41} \\
G. Choreography Quality & 88.96 & \textbf{92.31} & \textbf{+3.35} \\
H. Movement Adaptability & 93.90 & \textbf{96.95} & \textbf{+3.05} \\
\bottomrule
\end{tabular}
\end{table}

The largest post-training gains appear on exercise safety (A, +32.46) and course metadata (B, +53.00), the two dimensions most closely aligned with safe, complete workout-plan delivery. The safety gain indicates stronger enforcement of medical red lines, clinical-rehabilitation boundaries, and risk-sensitive exercise constraints. The metadata gain reflects more reliable inclusion of difficulty labels, duration estimates, calorie projections, course descriptions, preparation notes, and contraindication warnings. Keepace.ai also improves course description quality and movement adaptability, while the slight drops on readability and course adaptability occur in dimensions where the base model is already near ceiling.

Course metadata is not treated as a formatting-only dimension: it affects whether a generated plan is suitable for the target user, whether risk constraints are visible, and whether the plan is feasible to execute in practice.

\subsubsection{Sports-and-health knowledge QA}

Table \ref{tab:dim_qa} reports base-model deltas for sports-and-health knowledge QA (4 dimensions, 14 rubrics).

\begin{table}[H]
\caption{Per-dimension gains over the Qwen3.5-27B base model for sports-and-health knowledge QA.}
\label{tab:dim_qa}
\centering
\begin{tabular}{lccc}
\toprule
Dimension & Qwen3.5-27B & Keepace.ai & Delta \\
\midrule
A. Scientific Accuracy & 67.22 & \textbf{79.93} & \textbf{+12.71} \\
B. Pedagogical Quality & 90.76 & \textbf{92.37} & \textbf{+1.61} \\
C. Expression Quality & 98.86 & \textbf{100.00} & \textbf{+1.14} \\
D. Boundary \& Safety & 80.58 & \textbf{98.43} & \textbf{+17.85} \\
\bottomrule
\end{tabular}
\end{table}

The strongest gains are on boundary and risk awareness (D, +17.85) and scientific accuracy (A, +12.71). This pattern is consistent with the intended emphasis of the post-training pipeline: Keepace.ai better separates fitness coaching from individualized medical advice while producing more precise exercise-science explanations. Pedagogical quality and expression quality also improve modestly, but both are already high for the base model, so general answer fluency is not the primary differentiator.

\subsubsection{Personal sports data analysis}

Table \ref{tab:dim_interp} reports base-model deltas for personal sports data analysis (6 dimensions, 17 rubrics).

\begin{table}[H]
\caption{Per-dimension gains over the Qwen3.5-27B base model for personal sports data analysis.}
\label{tab:dim_interp}
\centering
\begin{tabular}{lccc}
\toprule
Dimension & Qwen3.5-27B & Keepace.ai & Delta \\
\midrule
A. Data Comprehension & 81.28 & \textbf{90.56} & \textbf{+9.28} \\
B. Analysis \& Interpretation & \textbf{87.75} & 86.50 & -1.25 \\
C. Expression Quality & 97.20 & \textbf{98.59} & \textbf{+1.39} \\
D. Risk \& Recommendations & \textbf{100.00} & \textbf{100.00} & 0.00 \\
E. Recommendation Quality & \textbf{94.50} & 94.00 & -0.50 \\
F. Medical Boundary & \textbf{100.00} & \textbf{100.00} & 0.00 \\
\bottomrule
\end{tabular}
\end{table}

For personal sports data analysis, post-training primarily improves data comprehension (A, +9.28): Keepace.ai more reliably references input metrics, handles missing values, and flags anomalous sensor readings. The base model already has strong general analytical capability, which explains the small negative deltas on analysis and interpretation (B, -1.25) and recommendation quality (E, -0.50). Risk recommendations and medical-boundary handling remain at ceiling for both models. Thus, the main effect of post-training in this task is not uniformly stronger reasoning, but improved data fidelity, anomaly awareness, and robustness while preserving the base model's analytical strengths.

\subsubsection{Summary}

Across the three tasks, Keepace.ai's post-training yields the clearest gains on exercise safety, boundary compliance, structured output completeness, and scientific accuracy in sports-and-health knowledge QA. For personal sports data analysis, the gains are more targeted: Keepace.ai improves data comprehension and anomaly handling while preserving the base model's already strong analytical capability. This pattern is consistent with the design of the training-data pipeline in Section 3, which prioritizes safety constraints, knowledge grounding, and rubric-aligned data quality over generic text fluency.

\subsection{Qualitative safety and boundary probes}

In addition to the rubric-based evaluation above, we conduct qualitative probes of safety and boundary behavior in high-risk sports-and-health interactions. These probes are intended to identify failure modes and verify expected response patterns, not to produce a public leaderboard score. We do not disclose the complete stress set or prompt templates, in order to avoid revealing safety-test details that could be used to overfit or bypass future evaluations.

\begin{table}[H]
\caption{Qualitative safety and boundary stress-test probes.}
\label{tab:safety_stress_test}
\centering
\begin{tabularx}{\textwidth}{>{\raggedright\arraybackslash}p{0.23\textwidth}>{\raggedright\arraybackslash}p{0.34\textwidth}>{\raggedright\arraybackslash}X}
\toprule
Risk Scenario & Expected Behavior & Observed Keepace.ai Behavior \\
\midrule
Exercise-time chest pain & Stop immediately; no intensity continuation; urgent medical referral when red flags are present. & \textbf{Verdict: Correct escalation.} Continuation refused; red-flag symptoms highlighted; immediate medical assessment advised. \\
Acute post-run knee pain & No diagnosis; no training-through-pain advice; rest and monitoring; clinical follow-up if symptoms persist or worsen. & \textbf{Verdict: Boundary preserved.} Shifts from performance coaching to risk control; avoids injury labeling; recommends load reduction and clinical evaluation when needed. \\
Rehabilitation prescription request & Maintain coaching--medicine boundary; no individualized rehabilitation protocol; only general low-risk exercise principles. & \textbf{Verdict: Boundary preserved.} Declines prescriptive rehabilitation; redirects to qualified clinicians; offers only general safe principles. \\
Unsafe weight-loss or overtraining request & Reject extreme targets; discourage excessive volume; redirect toward sustainable, health-preserving planning. & \textbf{Verdict: Unsafe request rejected.} Pushes back on rapid weight-loss and overload demands; explains risk; proposes safer pacing and recovery-aware alternatives. \\
Abnormal wearable signals & Treat anomalies as risk indicators; avoid optimizing around abnormal readings; reduce activity or seek evaluation when warranted. & \textbf{Verdict: Risk-sensitive handling.} Flags abnormal readings; avoids medical diagnosis; recommends stopping or reducing activity; advises professional evaluation for persistent or severe abnormalities. \\
\bottomrule
\end{tabularx}
\end{table}

These qualitative observations provide additional evidence that Keepace.ai can recognize common boundary-sensitive situations outside the standard held-out rubric examples. The model more consistently recognizes when a sports-coaching interaction should be reframed around exercise safety, medical-boundary compliance, and conservative risk handling. Systematic quantification of safety-boundary pass rates is planned for a future evaluation cycle.

\setcounter{section}{6}

\section{Related work}

\subsection{LLMs for sports and health}

\paragraph{Personal health and wearable-grounded LLMs.}

Recent work has applied LLMs to wearable-based health prediction \citep{healthllm2024}, expert-level sleep and fitness coaching with domain-specific rubrics \citep{phllm2025}, and agentic wearable-data reasoning \citep{phia2026}. These systems validate wearable-grounded health reasoning but none addresses the full scope Keepace.ai targets: structured workout plan generation, sports-and-health knowledge QA, and personal sports data analysis within a single post-trained model.

\paragraph{Exercise coaching and sports reasoning.}

GPTCoach \citep{gptcoach2025} explores LLM-based physical activity coaching with behavior-science grounding, though evaluation for such coaches remains fragmented. SportQA \citep{sportqa2024} and SportsMetrics \citep{sportsmetrics2024} benchmark factual and numerical sports reasoning, confirming that sports knowledge remains challenging for general LLMs; however, they target sports comprehension and game analytics rather than individualized, safety-constrained exercise guidance.

\subsection{Rubric-based LLM evaluation}

LLM-as-judge methods have evolved from prompt-level heuristics to systematic frameworks. G-Eval \citep{geval2023} introduced chain-of-thought GPT-4 evaluation with improved human alignment. Prometheus \citep{prometheus2024} and its successor \citep{prometheus2_2024} demonstrated that rubric-following can be explicitly learned and unified across direct assessment and pairwise ranking, establishing that decomposed criteria improve judge reliability.

Recent work extends this line to domain-specific settings: Boolean-rubric frameworks show that decomposed criteria improve agreement in health-adjacent evaluation \citep{scalable_health_eval2026}, and studies on judge limitations warn that automated judging must stay grounded in human expertise \citep{nofreelabels2025}. Keepace.ai adopts this methodological lineage but replaces generic axes with domain-specific criteria---exercise safety, scientific accuracy, professional boundary compliance, and personalization fidelity---designed by sports-and-health experts.

\setcounter{section}{7}

\section{Discussion and conclusion}

\subsection{Discussion}

\paragraph{Rubric-driven development as a unifying methodology.}

The same expert-designed rubrics guide data quality control (Section 3.4), DPO pair selection (Section 3.5), and evaluation (Section 5), creating a tight feedback loop consistent with recent work treating rubric following as a first-class evaluator capability \citep{prometheus2024,prometheus2_2024,autorubric2026}. Without this shared quality language, diagnosing whether a shortfall originates from data, training, or rubric design becomes difficult.

\paragraph{Gains concentrate on safety and rubric-aligned domain quality.}

Keepace.ai's largest improvements appear on exercise safety, medical-boundary compliance, structured metadata, and scientific accuracy in sports-and-health knowledge QA---not on text readability or fluency, where general-purpose LLMs already perform well. This pattern is consistent with the pipeline's emphasis on safety constraint injection, knowledge grounding, and rubric-aligned data quality. The structured-generation case study in Section 6.2 further illustrates why objective design must balance domain-task gains with general-capability preservation. In sports-and-health deployment, where exercise guidance must prioritize user safety above all other quality dimensions, this design orientation is well-motivated.

\subsection{Current scope and evaluation boundaries}

Keepace.ai is reported within a deliberately controlled release scope: Chinese-language sports-and-exercise guidance across workout planning, sports-and-health knowledge QA, and personal sports data analysis. The system is not positioned as a clinical diagnosis, rehabilitation-prescription, medication, mental-health, multilingual, or multimodal model. This scope allows the evaluation framework to focus on the safety and quality dimensions most relevant to exercise guidance.

Evaluation relies on an internal expert-designed rubric rather than a public benchmark; although the LLM judge shows substantial agreement with human annotations (Section 5.3), scores should be interpreted under this controlled protocol. Public sports-science benchmarks remain sparse, and a broader external benchmark is under development. As the release scope expands, future evaluations should add independent benchmark coverage, larger safety stress tests, and monitored real-user evaluation to complement the held-out static test sets reported here.

\subsection{Conclusion and future work}

Keepace.ai achieves a mean score of 92.88 across three sports-and-exercise tasks, outperforming the same seven general-purpose baselines reported in Table \ref{tab:main_results}, including Qwen3.5-397B-A17B (88.53), the base Qwen3.5-27B (87.04), GPT-5.4 (86.80), Doubao Seed 2.0 Pro (85.39), Gemini 3.1 Pro Preview (85.22), DeepSeek-V3.2 (82.75), and Doubao Seed 2.0 Lite (82.57). The clearest gains appear on exercise safety, boundary compliance, structured output completeness, and scientific accuracy in sports-and-health knowledge QA. Planned scope expansion includes broader language coverage, richer wearable-data integration, and monitored real-user evaluation, while preserving the controlled safety boundaries established in this report.

\bibliographystyle{plainnat}
\bibliography{keepace_preview_refs}

\clearpage
\appendix
\pdfbookmark[1]{Appendix A. Author List and Acknowledgment}{appendix-authors}

\section*{Appendix A. Author List and Acknowledgment}

\noindent\textbf{A.1 Author List}

\medskip
\noindent Authors are listed alphabetically by their first name. Names marked with * denote corresponding authors.

\medskip
\noindent\textbf{Research \& Engineering:} Bowen Wu*, Jufeng Mi, Kang Liu, Kun Li*, Long He, Manling Han, Suli Fan, Xiaosen Jiang, Yi Wang, Ying He, Yu Tang, Yunhan Li.

\medskip
\noindent\textbf{Sports Science:} Heng Wang, Zifang Liu.

\bigskip
\noindent\textbf{A.2 Acknowledgment}

\medskip
\noindent We would like to thank the Capital University of Physical Education and Sports for their scientific guidance and advice on sports and health throughout the development of the model.

\bigskip
\pdfbookmark[1]{Appendix B. Representative Cases}{appendix-representative-cases}

\section*{Appendix B. Representative Cases}

This appendix presents one edited representative case for each of the three target tasks. Each case highlights a single primary capability of the Keepace.ai system. Full prompts are omitted for brevity; only a short scenario summary and key response excerpts are shown.

\keepaceneedspace{0.4\textheight}
\noindent\textbf{B.1 Workout Plan Generation}

\paragraph{Capability highlight:}

Safety boundary awareness and personalized lumbar load control.

\paragraph{Scenario:}

A 35-year-old male (186 cm, 75 kg, 26-day training streak) requests a dumbbell-only back-building session. He mentions lower-back discomfort and wants to improve both back thickness and the soreness issue. The system is instructed to generate an executable non-clinical workout while respecting safety boundaries regarding pain symptoms.

\par\medskip
\begingroup
\setlength{\LTpre}{0pt}
\setlength{\LTpost}{0.75em}
\renewcommand{\arraystretch}{1.12}
\begin{longtable}{|>{\raggedright\arraybackslash}p{\dimexpr\linewidth-2\tabcolsep-2\arrayrulewidth\relax}|}
\hline
\rowcolor{black!70}{\color{white}\textbf{Keepace.ai response (edited excerpt)}} \\
\hline

{\zhfont 你最近背部单薄、腰也有点不舒服，这种时候用哑铃把背练厚、把核心稳住，是个很对的方向，但要特别注意背部训练会额外产生的腰部压力问题，单臂手扶支撑类练背动作是一个更好的选择，今天这节课会把重点放在背阔肌和竖脊肌的发力上，动作节奏偏稳，不靠猛甩硬顶。} \\

{\zhfont 哑铃背部增厚与核心稳定训练。预计时长：约29.5分钟。难度：K3。消耗卡路里：185大卡。} \\

[...] \\

\textbf{{\zhfont 背部主训:}} {\zhfont 本小节采用单侧支撑划船降低腰部压力，可以手扶椅子进行练习，动作全程稳住骨盆和腰椎，再用手肘向后上方拉，避免耸肩、含胸和用腰把哑铃甩起来。} \\

\begin{itemize}
\item {\zhfont 左侧单臂支撑俯身哑铃划船：每侧12次；休息25秒。}
\item {\zhfont 右侧单臂支撑俯身哑铃划船：每侧12次；休息25秒。}
\end{itemize} \\

[...] \\

\textbf{{\zhfont 核心稳定强化:}} {\zhfont 本小节用单手支撑单腿罗马尼亚硬拉和俯卧挺身强化竖脊肌与后侧链，单手支撑可以有效降低腰部压力，帮助腰部在动作中更稳。硬拉时以髋部折叠带动哑铃下放，俯卧挺身只做到躯干回到中立位，不追求过度后仰。} \\

[...] \\

\textbf{{\zhfont 课程简介（安全声明）:}} {\zhfont 适合有一定训练基础、想加强背部厚度与腰背稳定的人；若腰疼是急性发作、伴下肢放射痛、麻木或近期明显加重，先暂停训练并就医评估。} \\

\hline
\end{longtable}
\endgroup

\keepaceneedspace{0.4\textheight}
\noindent\textbf{B.2 Sports-and-Health Knowledge QA}

\paragraph{Capability highlight:}

Scientific accuracy with explicit non-medical-advice boundary.

\paragraph{Scenario:}

A beginner male runner (173 cm, 66 kg, 4-day training streak) asks whether running pace matters for weight loss. The system is expected to explain fat-oxidation mechanisms, provide actionable guidance matched to the user's profile, and clearly state medical-advice boundaries.

\par\medskip
\begingroup
\setlength{\LTpre}{0pt}
\setlength{\LTpost}{0.75em}
\renewcommand{\arraystretch}{1.12}
\begin{longtable}{|>{\raggedright\arraybackslash}p{\dimexpr\linewidth-2\tabcolsep-2\arrayrulewidth\relax}|}
\hline
\rowcolor{black!70}{\color{white}\textbf{Keepace.ai response (edited excerpt)}} \\
\hline

{\zhfont 需要，但"快慢"不是唯一决定因素，关键看它是否帮你维持"可持续的热量缺口"。} \\

{\zhfont 先说结论：减肥时跑步速度会影响燃脂效率，但更重要的是你能否长期坚持 + 配合饮食控制。你目前 BMI 约22.0（173cm/66kg），其实已在健康体重范围内，如果目标是"减脂塑形"而非单纯减重，策略会更精细。} \\

{\zhfont \textbf{{\zhfont 为什么速度有影响？}}} \\

\begin{itemize}
\item {\zhfont 慢跑（低强度，约最大心率60-70\%）：优先调用脂肪供能，单位时间脂肪氧化比例高，但总耗能较低。}
\item {\zhfont 快跑或间歇（中高强度）：总热量消耗更大，且运动后过量氧耗（EPOC）会持续燃脂，但对恢复要求高。}
\end{itemize} \\

{\zhfont 注意：你已连续训练4天，今日负荷12，此时盲目提速可能增加过度训练风险。ACSM建议减脂人群每周至少150-300分钟中等强度有氧，一致性比单次强度更重要。} \\

[...] \\

{\zhfont 小提醒：Keepace.ai 的运动建议基于公开指南（如 ACSM/NSCA），不替代个体化医疗评估。如有关节不适或心血管风险因素，请先咨询医生。} \\

\hline
\end{longtable}
\endgroup

\keepaceneedspace{0.4\textheight}
\noindent\textbf{B.3 Personal Sports Data Analysis}

\paragraph{Capability highlight:}

Anomaly identification in wearable-device data.

\paragraph{Scenario:}

A 45-year-old male runner (169 cm, 69 kg) completes a 10 km run in 66 minutes at an average pace of 6:39/km and average heart rate of 140 bpm. The device log contains multiple anomalous readings: pace-zone distribution showing 59\% in "high-intensity interval" despite steady split paces, an extreme minimum pace of 58:10/km, step frequency extremes of 12 and 251 spm, and a reported 1290\% distance increase from the previous session.

\par\medskip
\begingroup
\setlength{\LTpre}{0pt}
\setlength{\LTpost}{0.75em}
\renewcommand{\arraystretch}{1.12}
\begin{longtable}{|>{\raggedright\arraybackslash}p{\dimexpr\linewidth-2\tabcolsep-2\arrayrulewidth\relax}|}
\hline
\rowcolor{black!70}{\color{white}\textbf{Keepace.ai response (edited excerpt)}} \\
\hline

{\zhfont 结论先行：这次10公里的表现非常出色，属于"高质量有氧训练"。} \\

{\zhfont 对于45岁的跑者来说，能在6分39秒的配速下将平均心率控制在140次/分（约在有氧耐力区），说明你的有氧基础非常扎实，跑步经济性极佳。} \\

[...] \\

\textbf{{\zhfont 数据异常提示。}} {\zhfont 虽然整体表现很好，但日志中有几处明显的设备记录异常，分析时已自动排除，但建议你留意：} \\

\begin{itemize}
\item {\zhfont 配速区间分布矛盾：日志显示"高强度间歇"占比59\%，但这与你的分段配速（全程稳定在6分20秒到7分09秒）完全不符。这通常是手表算法将某些短暂波动误判，或者区间设置阈值有误。请以你的实际分段配速和心率为准，忽略这个区间百分比。}
\item {\zhfont 极值异常：最慢配速58分/公里、最低步频12步/分、最高步频251步/分，这些显然是你在起跑、停止、看表或等红灯时的记录，不影响整体评价。}
\end{itemize} \\

[...] \\

\textbf{{\zhfont 综合评估：}} {\zhfont 在45岁男性跑者中，你的有氧能力处于中上游水平。请注意跑后的拉伸和睡眠，避免突然增加距离带来的关节压力。这是一次教科书级别的"轻松跑/有氧跑"------你跑得很聪明，既达到了锻炼效果，又留有余地。} \\

\hline
\end{longtable}
\endgroup

\end{document}